\documentclass[preprint,12pt]{elsarticle}

\usepackage{multirow}
\usepackage{amssymb}
\usepackage{amsmath}
\usepackage[utf8]{inputenc}
\usepackage{newunicodechar}
\usepackage{subcaption}
\newunicodechar{├}{|}
\newunicodechar{└}{|}

\journal{Composites, Part A}

\begin{document}

\begin{frontmatter}



\title{A Process-Informed, Photorealistic Digital Twin for Vision-Based Inline Inspection of C-SMC Strip Deposition} 


\author{Patrick Flore} 

\affiliation{organization={Leibniz-Institut f\"ur Verbundwerkstoffe GmbH},
            addressline={Erwin-Schr\"odinger Stra{\ss}e, Geb. 58}, 
            city={Kaiserslautern},
            postcode={67663}, 
            state={},
            country={Germany}}

\author{Fabian R\"oder} 

\affiliation{organization={Leibniz-Institut f\"ur Verbundwerkstoffe GmbH},
            addressline={Erwin-Schr\"odinger Stra{\ss}e, Geb. 58}, 
            city={Kaiserslautern},
            postcode={67663}, 
            state={},
            country={Germany}}

\author{Kevin Chen} 

\affiliation{organization={Leibniz-Institut f\"ur Verbundwerkstoffe GmbH},
            addressline={Erwin-Schr\"odinger Stra{\ss}e, Geb. 58}, 
            city={Kaiserslautern},
            postcode={67663}, 
            state={},
            country={Germany}}

\author{Andreas Gebhard} 

\affiliation{organization={Leibniz-Institut f\"ur Verbundwerkstoffe GmbH},
            addressline={Erwin-Schr\"odinger Stra{\ss}e, Geb. 58}, 
            city={Kaiserslautern},
            postcode={67663}, 
            state={},
            country={Germany}}

\begin{abstract}
Inline quality assurance for Sheet Molding Compound (SMC) manufacturing is hindered by the lack of large-scale, accurately labeled image datasets for training machine-learning models. This work presents a process-informed, photorealistic digital twin for carbon-fiber SMC strip deposition. A rigid-body simulation reproduces the deposition kinematics (settling, friction, belt motion), while a physically grounded polarization-rendering model produces realistic appearance, so that every image carries exact, mutually consistent labels in synchronized formats: oriented bounding boxes, point annotations, density maps, and polarization renderings (Angle and Degree of Linear Polarization). Because the joint distribution of many overlapping strips cannot be specified analytically, simulation yields exact, scalable ground truth that manual annotation cannot provide. The twin produces over $16{,}000$ images across $3{,}300$ frames, and the synthetic data replicate near-isotropic fiber orientation (axial concentration $r < 0.05$) consistent with as-deposited commercial SMC. Trained purely on synthetic data, YOLOv11 and CSRNet reach detection $\text{mAP}_{50}$ up to $0.91$, with a dedicated oriented-bounding-box detector confirming the oriented labels are directly trainable ($\text{mAP}_{50} = 0.83$); on real manufacturing images, the synthetic-to-real difference in edge and texture structure (Kolmogorov--Smirnov distance $0.46$, distribution overlap $0.56$) is of the same magnitude as the difference between two real acquisition series of the same material ($0.41$, overlap $0.57$). The approach substantially reduces the labeling burden for camera-based inline inspection of fiber orientation and volume content.
\end{abstract}

\begin{keyword}
Sheet Molding Compound \sep Synthetic Training Data \sep Polarization Imaging \sep Digital Twin
\end{keyword}



\end{frontmatter}



\section{Introduction}
\label{section:introduction}
For decades, the global emphasis on reducing carbon dioxide emissions and improving energy efficiency has driven the transportation sector to shift toward lightweight construction. This long-term effort has established fiber-reinforced polymers (FRPs) as key alternatives to traditional materials like metals. Among FRPs, Sheet Molding Compound (SMC) parts are particularly attractive because they offer a balance of high specific stiffness and strength, low cost, and efficient manufacturability. SMC is a prepreg sheet: a semi-finished fiber-reinforced thermoset polymer matrix product that typically consists of a thermosetting resin (such as unsaturated polyester, vinyl ester, or epoxy), particulate fillers (e.g., calcium carbonate), and long, discontinuous fiber reinforcement (typically 25–50 mm in length).

The production of an SMC part begins with the creation of the semi-finished prepreg sheet on a compounding line, where the thermoset resin paste, fillers, and long chopped fibers or fiber bundles are combined and compacted onto carrier foils. The sheet then undergoes a maturing phase, which increases its viscosity to the level required for subsequent handling. After maturation, the SMC sheet is cut to the required size and, if necessary, preformed before being compression-molded into the final component.

The mechanical performance of an SMC component—particularly its strength and stiffness—is governed directly by the microstructure of the semi-finished sheet. This microstructure is described by two spatially varying fields: the fiber orientation distribution (FOD) and the fiber volume content (FVC). The FOD is a tensor-valued field, typically represented by a second- or higher-order orientation tensor that characterizes the local directional probability distribution of fibers or fiber bundles \cite{gorthofer_virtual_2019, sabiston_method_2019}. The FVC, in contrast, is a scalar field specifying the local fiber volume fraction within the composite \cite{castro_numerical_2025}. Together, these fields determine the local anisotropic mechanical behavior through established micromechanical homogenization relations \cite{anagnostou_hierarchical_2018,falaschetti_experimental_2025} and serve as the primary predictors of stiffness, strength, and failure-relevant properties at the component scale. However, local deviations from the expected orientation distribution or from the nominal fiber volume content inevitably arise due to the stochastic nature of fiber placement during sheet assembly. Additionally, it is still subject to debate whether a stochastically perfect process would yield a planar quasi-isotropic FOD \cite{meyer_direct_2020, feraboli_characterization_2009} or whether the forward feed in the direction of production would result in a non-isotropic FOD \cite{duhovic_quantitative_2025, nony-davadie_mechanical_2019}. 

Still, quality assurance during SMC manufacturing needs to define boundaries for possible deviations from the expected FOD and FVC and to provide efficient and precise means for measuring both. 
Therefore, in-line sensing of FOD and FVC during SMC sheet production is of great interest for quality assurance. Such an inline system must, however, satisfy several practical constraints. It must operate at the line speeds typical of SMC production, on the order of a few cm/s (the production line studied in this work runs at 2 cm/s), implying real-time data acquisition with minimal processing latency \cite{orgeas_sheet_2012}. It also requires sufficient spatial and angular resolution to detect deviations in fiber architecture, while remaining robust to surface variations, occlusions, and the inherent heterogeneity of discontinuous-fiber materials. A wide range of non‐destructive testing (NDT) methods has been proposed, each with different spatial/temporal resolution, accuracy, and practicality. These span X-ray computed tomography and scattering \cite{garcea_x-ray_2018, sugimoto_estimation_2022}, eddy-current sensing \cite{bardl_automated_2016, de_goeje_non-destructive_1992}, active infrared thermography \cite{fernandes_active_2015}, and optical microscopy \cite{eberhardt_fibre-orientation_2001}, alongside quality-control approaches specific to SMC production \cite{kraemer_quality_2014}.

While radiographic or tomographic methods can access bulk information \cite{garcea_x-ray_2018}, optical inspection approaches have emerged as attractive alternatives due to their non-contact nature and high acquisition rates \cite{meister_imaging_2021, atkinson_precision_2021, duhovic_quantitative_2025}. However, conventional optical approaches often require an extensive post-processing \cite{dobrich_machine_2023}, which limits their suitability for inline use. Polarization imaging therefore represents a particularly promising alternative to conventional optical methods for carbon-fiber-reinforced materials, as carbon fibers induce a pronounced polarization state in reflected light \cite{freitag_polarization_2014}. This enables fiber-orientation information to be extracted with substantially reduced post-processing effort, making the approach inherently better suited for inline applications \cite{schommer_polarization_2023}.

Machine-vision cameras with on-chip polarizers can image the angle of linear polarization (AoLP) of light from carbon fibers, providing pixel-level surface fiber orientations. These systems are inexpensive (1–3 kEUR) and operate at video rates, enabling real-time inspection. For uni- and bidirectional C-SMC laminates, polarization-based orientation measurements have been validated with an angular accuracy on the order of 1° \cite{atkinson_precision_2021, chiominto_using_2024}. Surface and bulk fiber orientations often correlate in C-SMC \cite{duhovic_quantitative_2025}, although the strength of this correspondence can depend on the production process. The method does not work for glass-fiber SMC, however, as glass fibers do not polarize light significantly. Nonetheless there are approaches to capture the drop-down area to gain more depth information of the SMC-Prepreg \cite{duhovic_3d-digitalization_2024}. Furthermore, the method does not perform explicit object recognition and therefore remains sensitive to occlusions, surface curvature and local texture variations, which can bias the AoLP signal and cause certain regions to dominate the orientation estimate. While highly effective for surfaces with clear carbon-fiber texture, its robustness decreases for multidirectional or irregular SMC materials, where overlapping fiber families and varying reflection properties can lead to averaged or unstable orientation fields \cite{duhovic_quantitative_2025}. Post-processing of the polarization images can mitigate these effects, but such steps reduce the method’s potential for inline application due to increased computational effort and latency. Pre-trained machine-learning models offer a possible alternative, yet they require large amounts of accurately labeled ground-truth data \cite{hutten_deep_2024} — data that are difficult or impossible to obtain experimentally during inline SMC production. Synthetic microstructure models therefore appear as a suitable source of labeled data.

Digital models for SMC prepregs have been proposed, typically generating microstructures by distributing fiber chips (pseudo-)randomly according to prescribed density functions \cite{teuwsen_direct_2020, chen_multiscale_2018}. While suitable for statistical analyses, such models do not capture process- or appearance-related variability such as surface topology and local thickness variations, illumination effects, and deposition dynamics. Such factors, however, are known to be essential for training machine-learning models in image-based inspection tasks, as ML methods require realistic variations in texture and lighting to generalize effectively \cite{tremblay_training_2018}. To address these limitations, the present work introduces a simulation framework in which fiber chips are not placed directly but are dropped onto a moving conveyor belt under rigid-body dynamics, incorporating effects such as gravitational settling, friction, belt motion, and controllable lighting conditions. This enables the generation of realistic, fully annotated synthetic image datasets for supervised learning of fiber-orientation and fiber-volume-content estimation. The simulation framework can also be used to generate annotated images for conventional optical inspection approaches, thereby enabling benchmark datasets for both polarization-based and purely optical fiber-orientation analysis. The resulting digital representation of the SMC semi-finished material can be integrated into a virtual process chain—such as press forming simulations—to assess how the initial fiber architecture affects press parameters and the material properties of the final component.

\section{Research Questions and Approach}

Analyzing the current state of the art, the major obstacles in developing a real-time optical quality inspection system for SMC production are the lack of large-scale, accurately labeled datasets, the meso- and macro-scale variability of the fiber arrangement in multidirectional SMC sheets, and the need for computationally efficient inline methods. Therefore, this paper addresses the following research questions:
\begin{enumerate}
    \item How can large-scale, fully annotated image datasets of SMC semi-finished parts be generated \emph{in silico}, capturing realistic meso- and macro-scale fiber-arrangement variability, lighting, and surface effects?
    \item How can these datasets support the development and evaluation of camera-based, real-time quality assurance methods for SMC production?
    \item How can machine learning algorithms for fiber-orientation and fiber-volume-content estimation be trained and validated using such synthetic data?
\end{enumerate}

The research approach pursued in this work is the development of a scientific artefact in the form of a \emph{process-informed, photorealistic digital twin} of the SMC manufacturing process. Classical computational approaches for fiber-orientation estimation are often too slow for inline application; by contrast, machine learning-based inference can achieve real-time performance once trained. The digital twin models the deposition and arrangement of fiber chips on a moving conveyor, including gravitational settling, friction, belt motion, and controllable lighting conditions. By generating fully annotated synthetic images, it serves as a virtual testbed for both data-driven and conventional optical inspection methods, enabling systematic evaluation and training of inline quality assurance algorithms.

\section{SMC Surface Analysis of production material}

During the manufacturing of SMC material, there are several in-between states relevant for understanding material structure. As the SMC is generated by randomly distributing chopped fibers on a plastic carrier film, its initial randomness state gets set. Upon further compaction along the process, this state is matured further, until finally the material selection, positioning and pressing process achieves the final FOD and FVC. Among these steps in between roving chopping and finished part, the ideal position to initial FOD and FVC is the initial randomized distribution generated by the chopping and dropping of the fibers onto the plastic carrier film. Here, initial positions and orientations are set and could still be influenced. Additionally, the lack of top layer and the pre-compaction rectangular shape of rovings simplyfies machine vision based approaches. 

\subsection{SMC material production setup}

The images used as ground truth for the comparison to the digital twin generated in this work are produced on a SMC-line at the IVW. (Schmidt und Heinzmann SMC Line600) It can  produce up to 600 mm of material width whilst consuming up to 23 rovings in parallel to achieve areal weights of up to 8 $\text{kg/m}^2$. In addition to the common features of smc production lines such as carrier film, resin pase application, chopper, compactor and roller, the machine features an additional unwinding unit to introduce dry textiles into the process in between the chopping and application of the top layer. Due to the additional distance, an area with settled down fibers is created that used to capture the full-frame ground truth images used in this work. Here, the camera setup is placed with its additional lighting and fixtures. 

The images captured for this work are taken within a production run of a usual production-run SMC (Polynt SMCarbon 80, Polynt SPA, Italy), featuring epoxy resin applied with a squeegee box gap of 1 mm. The rovings used for production are 15k tows chopped to a nominal length of 25 mm and guided onto the production width of 360 mm using baffles. The rotation speed of the chopper of 35 rpm at the carrier film speed of 2 cm/s results in an average areal weight of roughly 1 $\text{kg/m}^2$. This results in a mostly closed surface of chopped carbon fibers, leaving the carrier film only sparsely uncovered. 

\subsection{Camera setup to capture ground-truth data}

Using the aforementioned position of the unwinding unit, the acquisition setup is placed right after the chopped fiber area. It consists of a lighting box ensuring proper light distribution within the camera frame by including LED lighting units (SBL-0130/SBL-0150, MBJ Imaging GmbH) and featuring protective, color filtering screens to allow external monitoring of the material passing through it at a lower level of lighting.  The lighting box also serves as a mounting frame for the polarization camera used to capture the images. Mounted at a height of 980 mm, using a fixed focal lens (HF16HA-1B, Fujifilm Holdings Corporation, Japan) the full-frame images fully capture the 360 mm width of the production line. 

The camera itself (VCXU-50MP, Baumer Electric AG, Switzerland) is an industrial usb camera for c-mount objectives featuring a monochrome sensor with a four directional on-chip polarizer (SONY IMX250MZR). The micro-wire grid polarizers are arranged directly on the chips surface as a two-by-two array similar to bayer color filter arrays as a Division-of-focal-plane sensor. It features the analyzer angles $0^\circ$, $45^\circ$, $90^\circ$ and $135^\circ$. This allows simultaneous acquisition of four polarization images in one shot, capturing polarization information in real time. 
The resulting raw image captured by the camera can accordingly be viewed as four simultaneously captured images for the four analyzer angles, with a slight offset. From these, according to stokes formulations, all linear polarization components (S0, S1, S2) of the original image can be calculated. For practical use with SMC, they are used to calculate two easily interpreted false-color images using the Degree of linear Polarization (DOLP) and the Angle of linear Polarization (AOLP). The DOLP image can be directly used to separate carbon fibers from unpolarized backgrounds, providing information about the FVC, whilst the AOLP directly represents the fibers' orientation and can thus be easily applied on the calculation of FOD. Among them, the AOLP can be used as the single most important image information, since it is only calculated for polarized sections of the image and thus already contains the essential information of the DOLP as well. It is accordingly the foremost focus of subjective comparison to the created synthetic image.

\section{A Process-Informed Digital Twin for SMC Production}


This section describes the created set of tools, concepts and workflows for generating synthetic data that is required for developing camera based-procedures for in-line quality assurance during SMC manufacturing. It includes the generation of a large number of roving strands with constant cross section but variable length, their dropping onto the moving conveyor belt and their physical interaction with the belt, which - together with some enforced variance during the spawning of the strips - results in a realistic settlement of the strips onto the belt. However, the simulation does not explicitly include
the resin, which is typically present on the conveyor belt when the fiber strips fall onto it. This is consistent with the situation during the development of camera-based inspection methods, which was performed on multiple series of images acquired directly within the chopping area. Furthermore, it is assumed that this has no significant impact on the final \emph{distribution} of
the location and orientation of the strips, at least not to an extends that defies the development of such inspection methods. The same applies to the non-inclusion
of the top carrier film which is applied to the sheet before it is spooled at the end of the SMC line.

\subsection{Simulation of the SMC Manufacturing Process}

The newly created simulation framework consists of three main components: 
(i) Blender 4.0 (Blender Foundation, Amsterdam, Netherlands), an open-source 3D computer graphics software package. Blender's integrated rigid body physics engine, based on the Bullet Physics library, is used for simulating realistic mechanical interactions between fiber strips and the conveyor belt.
(ii) A self-developed Python automation script (\texttt{stripGen\_Ana\_automate.py}) that orchestrates scene setup, physics baking, data analysis, and rendering.
(iii) An external runner script (\texttt{run\_simulation.py}) that manages batch processing and parallelization across multiple simulation seeds. The runner script takes all relevant simulation parameters from a JSON configuration file (\texttt{config.json}), including the number of strips to simulate, spawn timing intervals, total simulation duration and frames for density measurement.

\begin{figure}[t]
    \centering
    \includegraphics[width=\textwidth]{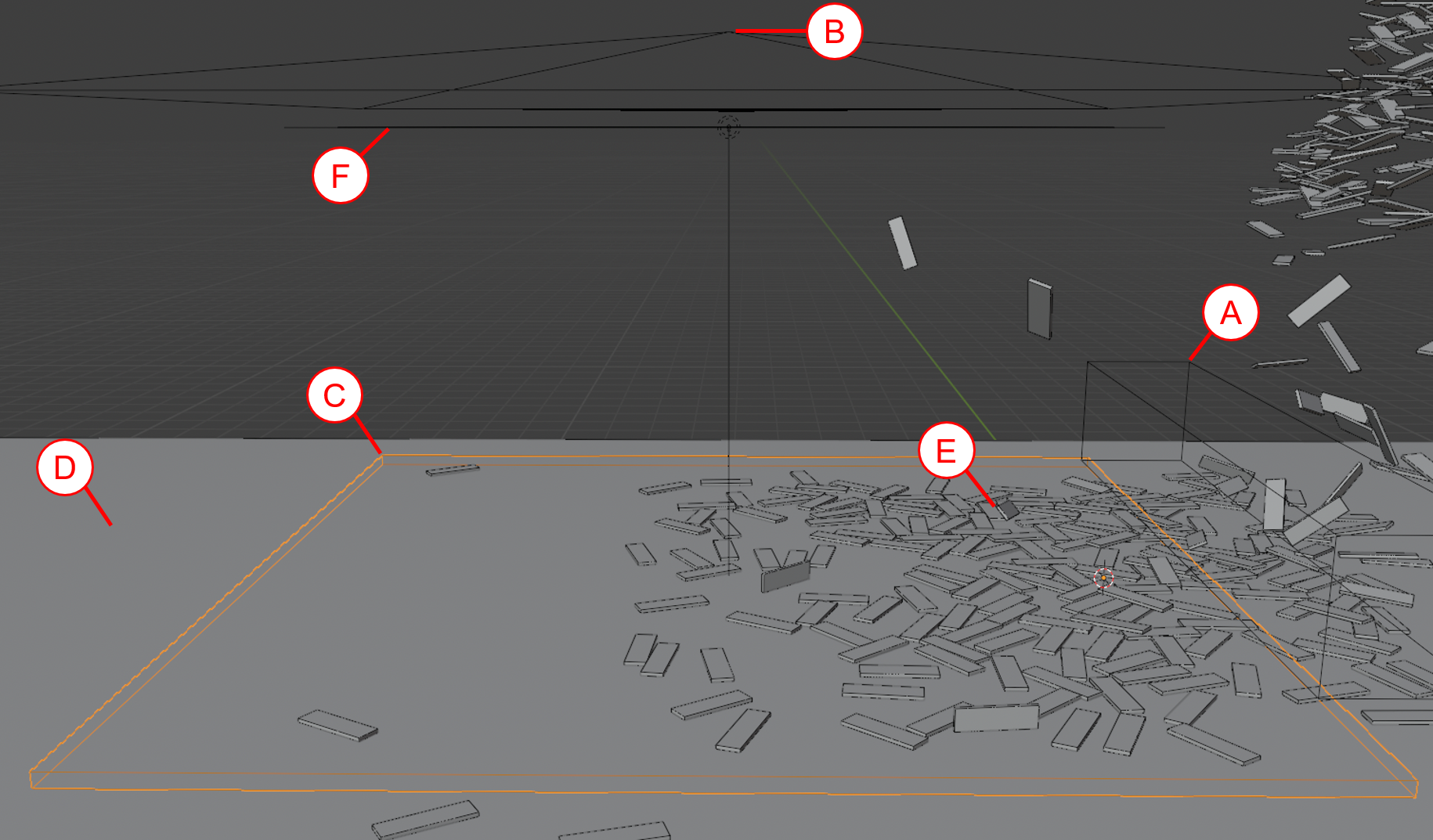}
    \caption{Simulation as displayed in Blender’s GUI. (A) shows the spawning region for the outlet, (B) indicates the camera’s origin point used for rendering, (C) marks the measurement box used for analysis and calculations, (D) represents the rigid-body conveyor-belt model, (E) shows individual fiber strips, here visibly stacked on top of one another and (F) shows the illumination area.}
    \label{fig:setup_annoted}
\end{figure}

The simulation framework replicates a typical SMC manufacturing line consisting of
a moving conveyor belt with configurable speed and friction properties,
 an outlet aperture positioned above the belt from which strips are spawned, and 
a measurement box for density quantification inside a world coordinate system whose x-axis is aligned with the belt direction of motion, the y-axis parallel to the belt width, and the z-axis representing the vertical direction, see Figure \ref{fig:setup_annoted}.

The simulation framework operates in two distinct phases. In the \textit{setup phase}, the automation script initializes the scene by clearing any existing objects, generating template geometries for strips, creating materials with appropriate visual and physical properties, and pre-computing spawn schedules to prevent initial geometric collisions. Subsequently, the rigid-body simulation is executed across all specified frames, with the rigid body dynamics system computing trajectories and contact forces between strips and with the conveyor belt. In the \textit{analysis phase}, executed after physics baking is complete, the framework advances to each specified measurement frame, computes both volumetric (3D) and projected (2D) density metrics within a defined measurement volume, renders polarization images for four distinct polarization angles, generates object detection annotations in YOLO (You Only Look Once) format, and persists all configuration parameters and computed metrics to an SQLite database for subsequent analysis.
This separation of setup and analysis phases, combined with deterministic random seed control, ensures that each simulation is fully reproducible while enabling efficient batch processing across parameter variations.

Individual fiber strips are represented as rectangular cuboid primitives with dimensions sampled from specified ranges.
Each strip is instantiated as an independent rigid body object with assigned physical properties. Mass is computed automatically from geometry and a specified material density. Surface interaction properties include a friction coefficient, which governs the sliding resistance between strips and during fiber-to-belt contacts, and a restitution coefficient controlling energy dissipation during collisions. These parameters are configurable model inputs, set to reproduce the qualitative deposition behavior of carbon-fiber strips observed in SMC manufacturing rather than measured as constitutive constants; their values are documented in the released code so the calibration can be inspected and adapted to other machines, chip geometries, and material systems.

The simulation deliberately models only the rigid-body geometry and kinematics of strip deposition, together with photorealistic appearance; it does not model the constitutive process physics of SMC, such as resin flow, viscosity, consolidation, or fiber-matrix mechanics. This scope is sufficient and appropriate for the intended purpose: the training targets used in this work (oriented bounding boxes, centroids, coverage density, and orientation maps) depend only on strip pose and overlap, which rigid-body deposition reproduces with exact ground truth, and not on constitutive material fidelity. The central advantage of generating this data by simulation is that the joint distribution of many overlapping strip placements, which is combinatorially infeasible to specify analytically, is produced directly, together with exact and scalable labels that manual annotation of real images cannot match.

A critical challenge in multi-object rigid-body simulation is avoiding initial geometric overlaps that lead to unrealistic collision forces and numerical instabilities. To address this, a deterministic spawn scheduling algorithm was developed that plans spawn times and positions to minimize initial collisions.

First, for each strip, the algorithm determines the planned spawn frame based on the spawn interval parameter and the strip index. It then pre-computes geometric properties using a seeded random number generator: the strip’s width is set to 0.3 units, its thickness to 0.05 units, and its length is sampled from $[1.0, 1.1]$ units. The spawn position has a fixed $x,z$ location, while the $y$-coordinate is sampled across the full width of the feeding aperture. Next, the algorithm randomizes the initial orientation to simulate natural tumbling during free fall by computing three Euler angles—yaw (around the vertical axis) from $[-5^\circ, 5^\circ]$, pitch from $[-10^\circ, 10^\circ]$, and roll from $[-10^\circ, 10^\circ]$—which provide sufficient variance while keeping the strip predominantly horizontal. Finally, it computes the centroid vector by calculating the 3D vector to the strip’s centroid.

Next, for each strip in ascending order of planned spawn frames, the following procedure is applied until all strips are spawned. The algorithm checks the spacing relative to previously spawned strips by computing the Euclidean distance between centroids. It then computes the minimum safe distance by summing (i) the maximum of the half-extents of both strips in $x, y, z$, (ii) an estimate of how far the earlier strip has already fallen, and (iii) a predefined minimum clearance (default 0.8 units). If the centroid distance exceeds this minimum safe distance, the strip is spawned; otherwise, its spawn is delayed to the next frame.

The combination of minimum clearance and progressively increasing falling distances for delayed strips effectively eliminates near-contact instabilities in the physics solver.

Each strip is spawned with zero initial velocity, subsequently acquiring motion through gravitational acceleration and aerodynamic effects implicitly captured in the physics engine's damping parameters. A small vertical offset (0.1 length units) is applied to ensure strips are positioned above the belt surface at spawn time. The Bullet physics engine then computes strip trajectories, taking into account gravitation and contact forces and torques when strips collide with the belt surface or with other strips, thereby replicating the industrial process of dropping chopped fiber strands onto the belt. 

The conveyor belt is modeled as a kinematic rigid body with predefined and constant linear velocity in the negative $x$-direction (default: $-0.01$ length units per frame). Unlike dynamic rigid bodies, kinematic bodies are not influenced by collision forces but exert forces on colliding objects. 
Therefore, the belt's motion imparts forward velocity to strips through friction without being slowed. Once strips settle onto the belt surface, they acquire horizontal velocity matching the belt speed while maintaining their relative positions and orientations. The simulation continues for a specified number of frames (default: 900 frames), allowing sufficient time for all strips to spawn, drop and settle into stable configurations.

\subsection{Density Calculation}

Quantifying fiber distribution is required for validating simulation and for generating ground truth labels. Two complementary density metrics were implemented: a volumetric density $\rho_\mathrm{3D}$ and a projected  areal density $\rho_\mathrm{2D}$ representing the coverage observed in top-down imaging.

$\rho_\mathrm{3D}$ is measured by placing a rectangular measurement volume (the so called measurement box) over the conveyor belt at a fixed location downstream from the outlet, ensuring that all strips have settled and are moving with the belt when measurements are taken.
A strip is considered to be inside the measurement box if at least one of its eight corners lies within the measurement box. For each of these strips, their full strip volume is included in the total volume of strip material within the measurement box $V_{\text{strips}}$, resulting in a slight overestimation of the actual density when strips intersect the box boundaries. 
$\rho_\mathrm{3D}$ is then the ratio of $V_{\text{strips}}$ and the volume of the measurement box $V_{\text{box}}$:
\begin{equation}
\rho_{\text{3D}} = \frac{V_{\text{strips}}}{V_{\text{box}}}
\end{equation}

Box size and position are the same in all simulation runs. Density measurements are performed at multiple specified frames during the simulation, capturing the temporal evolution of fiber packing as strips accumulate on the belt.

Analogously, $\rho_{\text{2D}}$ is defined as:
\begin{equation}
\rho_{\text{2D}} = \frac{A_{\text{shadow}}}{A_{\text{box}}}
\end{equation}
where $A_{\text{shadow}}$ is the total projected shadow area of all strips within the measurement box and $A_{\text{box}}$ is the horizontal cross-sectional area of the measurement volume. When computing $A_{\text{shadow}}$, overlapping regions of the projected strip areas are counted only once, so $\rho_{\mathrm{2D}}$ measures the coverage fraction rather than the total accumulated projected area.  In essence, this method of calculation results in an output identical to the use of the DoLP image to differentiate areas based on their polarizing attributes.

Density measurements are performed at multiple user-specified frames throughout the simulation, generating temporal profiles of fiber accumulation. These profiles characterize the transient behavior of the deposition process, including initial sparse coverage, gradual build-up, and eventual steady-state packing density.

\subsection{Polarization Imaging Simulation}

Polarization imaging is an emerging technique in industrial inspection that leverages the orientation-dependent optical properties of fibrous materials. To facilitate the development of polarization-based fiber orientation detection workflows, the proposed simulation framework includes a physically grounded polarization rendering system, thought to represent the typical hardware configuration of the used cameras, without the usual mosaicing, however.

Carbon fibers, when illuminated with unpolarized light reflect light that is generally partially linearly polarized, an effect stemming from the differences between the absorptivity of the fibers parallel and perpendicular to the to the fiber axis \cite{freitag_polarization_2014, atkinson_high-sensitivity_2018}. This effect is dependant on the material composition, its structure and the wavelength distribution of the light. To measure this effect using machine vision, polarization filters are implemented to create a difference in intensity on a light-sensitive component. They again reduce the intensity of the light by filtering the polarized component, following Malus' law:
\begin{equation}
I_{\mathrm{trans}}^{\mathrm{pol}} = I_{\mathrm{inc}}^{\mathrm{pol}} \cos^2(\phi - \phi_a),
\end{equation}
where \(\phi\) is the fiber’s principal axis orientation in the image plane, and \(\phi_a\) is the orientation of the analyzer.

Real filters neither achieve full transmission in parallel orientation nor perfect extinction perpendicular to the wave orientation. Additionally, strong relationships between the wavelength and the resulting extinction ratio can be expected and are present in the camera used. Since noise and distortion as well as challenges resulting from the Dofp-Sensor are out of scope for the current work, a simplified model directly representing the DOLP with a correction factor is used in the following equation:
\begin{equation}
\label{eq:dolp_correction}
I_{\mathrm{em}} = I_{\mathrm{pol}} + I_{\mathrm{unpol}}
= d \, I_{\mathrm{em}} + (1-d) \, I_{\mathrm{em}},
\end{equation}
where $d \in [0,1]$ is the degree of linear polarization (DoLP) of the emitted light.
As the unpolarized fraction is transmitted at half intensity, independent of angle:
\begin{equation}
I_{\mathrm{trans}}^{\mathrm{unpol}} = \frac{1}{2} I_{\mathrm{inc}}^{\mathrm{unpol}}.
\end{equation}
The total transmitted intensity is therefore:
\begin{align}
I_{\mathrm{trans}} &= I_{\mathrm{trans}}^{\mathrm{pol}} + I_{\mathrm{trans}}^{\mathrm{unpol}} \\
&= d \, I_{\mathrm{inc}} \cos^2(\phi - \phi_a) + \frac{1-d}{2} I_{\mathrm{inc}}.
\end{align}
Therefore, $d$ corresponds exactly to the DoLP, as can be verified from the standard contrast-based definition:
\begin{equation}
DoLP = \frac{I_{\mathrm{trans}}^{\max} - I_{\mathrm{trans}}^{\min}}{I_{\mathrm{trans}}^{\max} + I_{\mathrm{trans}}^{\min}} = d,
\end{equation}
with
\begin{equation}
I_{\mathrm{trans}}^{\max} = P I_{\mathrm{inc}} + \frac{1-P}{2} I_{\mathrm{inc}} = \frac{1+P}{2} I_{\mathrm{inc}}, \quad
I_{\mathrm{trans}}^{\min} = \frac{1-P}{2} I_{\mathrm{inc}}.
\end{equation}
Hence, $d \in [0,1]$ interpolates between completely unpolarized light ($d=0$) and fully linearly polarized light ($d=1$), enabling systematic investigation of partial polarization effects.

To simulate a division-of-focal-plane polarization camera or a rotating analyzer setup, the framework renders each measurement frame at analyzer orientations $0^\circ$, $45^\circ$, $90^\circ$ and $135^\circ$. These orientations correspond to standard configurations used in Stokes parameter measurement and are sufficient to reconstruct the full linear polarization state \cite{tyo_review_2006, powell_calibration_2013}.
For each analyzer angle, the shader's global analyzer angle parameter is updated across all materials in the scene, and a rendering pass is executed. 
The orthographic camera is positioned directly above the measurement box and aligned with its principal axes, ensuring a calibrated top-down view with known geometric correspondence between image pixels and physical coordinates on the belt surface.

From the four intensity images acquired at different analyzer orientations, the pixel-wise angle of linear polarization (AoLP) is computed using standard Stokes formalism \cite{tyo_review_2006}. The AoLP represents the dominant fiber orientation at each spatial location and is calculated as:
\begin{equation}
\text{AoLP} = \frac{1}{2} \arctan\left(\frac{I_{45^{\circ}} - I_{135^{\circ}}}{I_{0^{\circ}} - I_{90^{\circ}}}\right)
\end{equation}
where $I_{0^{\circ}}, I_{45^{\circ}}, I_{90^{\circ}}$ and $I_{135^{\circ}}$ denote the intensity measured with analyzer at angle $0^{\circ}$, $45^{\circ}$, $90^{\circ}$ and $135^{\circ}$ respectively. To avoid numerical instabilities from division by near-zero values, a small regularization constant $\epsilon = 10^{-6}$ is added to the denominator. The resulting AoLP values span the range $[-\pi/2, +\pi/2]$, corresponding to the full range of possible linear orientations. For visualization purposes, the AoLP values are normalized to the 8-bit grayscale range $[0, 255]$ through linear scaling:
\begin{equation}
\text{AoLP}_{\text{norm}} = \left\lfloor \frac{\text{AoLP} + \pi/2}{\pi} \times 255 \right\rfloor
\end{equation}

The computed AoLP images provide spatially resolved orientation maps that can serve as ground truth for training orientation-sensitive neural networks or for validating classical orientation detection algorithms. This synthetic polarization data generation capability closes a significant gap in available training datasets for the development of polarization-based industrial inspection systems \cite{hutten_deep_2024}. Due to the calculation of the AOLP being independent of the unpolarized components of the intensity, modeling only the polarized components within the render pipeline fulfills all the needs for generating synthetic AOLP images. For the same reason, the degree-of-linear-polarization correction factor $d$ (Eq.~\ref{eq:dolp_correction}) enters only the intensity and the DoLP-derived quantities, not the AoLP. Because the results reported here use the AoLP channel and add no sensor noise, the specific value of $d$ is at most numerically relevant to them.

The rendering model deliberately captures the first-order polarization physics (the Malus-law analyzer response and a per-material degree of linear polarization) while leaving several second-order appearance factors for later refinement. In real acquisitions, surface micro-roughness and fiber-texture detail, out-of-plane strip tilt, and the direction and diffuseness of the illumination all modulate the measured intensities, and hence the recovered AoLP, as does the wavelength dependence of the analyzer extinction ratio. In the present model the strip surfaces carry a simplified, spatially uniform appearance, the illumination is uniform, and only in-plane orientation is encoded, so that the rendered AoLP reflects strip orientation without these confounds.

Two appearance mechanisms deserve explicit mention. The macroscopic fiber orientation together with the surface texture can produce specular reflections that are themselves partially polarized, and the strongly anisotropic meso-scale structure of the deposited strips gives rise to direction-dependent diffuse reflection. In the comparison drawn in this work both the real acquisition and the simulation assume global, diffuse illumination, and the real frames are recorded under constant LED lighting with stable intensity and a fixed spectrum, so that neither reflection effect dominates the measured signal. The AoLP itself does not depend on the overall brightness; its residual variability on real data is instead driven by sensor noise and by interruptions in the fiber texture. A further effect left deliberately unmodelled is the polarization introduced at Brewster-angle reflections, which the present ray tracer cannot represent because it propagates no polarization state through the light transport, and which is therefore left to future work. The experimental comparison is moreover restricted to dry, as-deposited strips, so that resin-driven dissolution of the fiber bundles and wetting-induced surface changes do not occur; the appearance of a strip therefore no longer changes once it has settled on the belt, which keeps the number of appearance-governing factors low.

The controlled introduction of surface roughness, normal-mapped texture, out-of-plane orientation, and lighting diversity is therefore deferred to the planned refinements in Section~\ref{sec:outlook}, where it is expected to narrow the residual synthetic-to-real appearance gap quantified in Section~\ref{sec:real_world_transfer}.  

\subsection{Ground Truth Data Generation}

A primary motivation for developing this digital twin framework is to generate large-scale, accurately annotated training datasets for deep learning-based object detection and segmentation algorithms. The simulation environment provides perfect knowledge of object locations, geometries, and properties, enabling automatic generation of various ground truth annotation formats.

\subsubsection{YOLO Oriented Bounding Box Annotations}

The pipeline generates object-detection annotations in the oriented bounding box (OBB) format, compatible with the YOLO family of real-time detection networks \cite{redmon_you_2016, khanam_yolov11_2024}. Unlike an axis-aligned bounding box, an OBB additionally encodes the in-plane rotation of each object, so the box follows the true orientation of an elongated strip rather than only its horizontal and vertical extent. For each measurement frame, all strip objects are enumerated and their world-space vertices obtained by applying the object transformation matrices to the mesh vertices; a strip is annotated only if its vertical ($z$) span overlaps the measurement box, so that only strips within the sampled volume are labeled.

For each annotated strip, the label stores the box center, the projected length and width of the strip (the extents of its own long and short axes projected onto the belt plane), and the in-plane orientation angle of its long axis. All quantities are read directly from the simulation geometry, normalized to the measurement-box dimensions, and clamped to $[0, 1]$ for strips that intersect the box boundary. Each strip therefore yields one plain-text line $(\text{class}, x_c, y_c, \ell, w, \theta)$, with $\ell$ and $w$ the normalized oriented length and width and $\theta$ the orientation angle. Because every value is taken from the exact simulation state, the oriented boxes are exact and their angle varies per strip with the deposited orientation; there is no constant-angle or axis-aligned approximation (see Fig.~\ref{fig:parameter_control}C).

\subsubsection{Dense Ground Truth for Segmentation}

While bounding boxes provide sufficient supervision for object detection tasks, instance or semantic segmentation networks require pixel-level annotations. The simulation framework's perfect knowledge of object geometry enables generation of precise segmentation masks.

Although not explicitly implemented in the current pipeline, the framework architecture supports extension to dense annotation formats. Potential approaches include: (1) rendering object ID passes where each strip is assigned a unique color, enabling automatic extraction of instance masks; (2) rendering semantic label maps where all strips are assigned the same label value, distinguishing foreground (strip) from background (belt); or (3) rendering depth maps and surface normal maps to provide geometric cues for downstream processing.

\subsubsection{Orientation and Geometric Attributes}

Beyond spatial localization, the simulation framework has access to rich per-object attributes that can serve as additional supervision signals. For each strip, the available properties include its orientation, where the strip’s 3D rotation is known in Euler angles or quaternion form and its in-plane orientation angle can be extracted from the projection onto the measurement plane; its dimensions, with exact width, length, and height recorded to support regression tasks for size estimation; and its spawn metadata, which provides temporal information such as the planned spawn frame, the actual spawn frame, and the time the strip has spent in the scene, all of which offer useful context for tracking and temporal modeling.

These attributes can be incorporated into multi-task learning frameworks or used to generate auxiliary supervision signals. Whereas the in-plane orientation angle is already provided directly by the oriented bounding boxes described above (as used for rotated-object detection, common in aerial image analysis), the recorded per-strip dimensions and spawn metadata offer additional regression and temporal-tracking targets.

\subsubsection{Annotation File Organization}

The ground truth annotation system maintains a strict correspondence between rendered images and annotation files. Each frame generates four PNG images corresponding to analyzer angles 0°, 45°, 90°, 135°, a computed orientation map, and a text file containing oriented bounding box annotations for all visible strips.

All files share a common base filename encoding the run timestamp, seed, and frame number, enabling unambiguous association and batch processing. This organizational structure is compatible with standard deep learning training pipelines and facilitates dataset versioning and management.

\subsubsection{Point Annotations from Box Centroids}

For density estimation tasks, the oriented-box annotations are converted to point annotations representing object centroids. This conversion is performed post-hoc during dataset preprocessing. For each oriented bounding box $(\text{class}, x_{\text{center}}, y_{\text{center}}, \ell, w, \theta)$ in normalized coordinates, the pixel-space centroid is computed from its center as:
\begin{align}
x_{\text{point}} &= x_{\text{center}} \cdot W_{\text{img}} \\
y_{\text{point}} &= y_{\text{center}} \cdot H_{\text{img}}
\end{align}
where $W_{\text{img}}$ and $H_{\text{img}}$ are the image width and height in pixels.

These point annotations are stored as CSV files with columns \texttt{x, y}, maintaining the one-to-one correspondence between image files and annotation files. This representation is particularly suited for crowd counting and density estimation networks that operate on point-based supervision rather than bounding boxes.

\subsubsection{Gaussian Density Map Generation for CSRNet}
\label{sec:density_map_generation}

For training density estimation networks such as CSRNet (Congested Scene Recognition Network) \cite{li_csrnet_2018}, point annotations are transformed into continuous density maps through Gaussian kernel convolution. This process converts discrete object locations into a smooth spatial distribution that the network learns to predict.
The \emph{density map generation algorithm} proceeds as follows:
\begin{itemize}
\item \textbf{Binary Point Map Construction:} A binary map $M_{\text{binary}}$ of the same dimensions as the input image is initialized to zero. For each annotated point $(x_i, y_i)$, the corresponding pixel is set to unity:
\begin{equation}
M_{\text{binary}}(x_i, y_i) = 1 \quad \text{for all annotated points } i
\end{equation}
\item \textbf{Gaussian Convolution:} The binary point map is convolved with a 2D Gaussian kernel $G_\sigma$ with standard deviation $\sigma$:
\begin{equation}
D(x, y) = (M_{\text{binary}} * G_\sigma)(x, y) = \sum_{i,j} M_{\text{binary}}(i, j) \cdot G_\sigma(x-i, y-j)
\end{equation}
where 
\begin{equation}
G_\sigma(x, y) = \frac{1}{2\pi\sigma^2} \exp\left(-\frac{x^2 + y^2}{2\sigma^2}\right)
\end{equation}
The parameter $\sigma$ controls the spatial extent of influence for each object. In this work, a value of $\sigma = 4$ pixels was used as a baseline, representing a compromise between localization precision and numerical stability. The Gaussian filter is applied with constant boundary padding to avoid edge artifacts.
\item \textbf{Spatial Downsampling:} Deep convolutional architectures typically produce feature maps at reduced spatial resolution due to pooling and strided convolution operations. To match the network's output resolution, the density map is spatially downsampled by a factor corresponding to the output stride. For CSRNet, which produces outputs at $1/8$ of the input resolution, the downsampling is performed through block summation:
\begin{equation}
D_{\text{down}}[r, c] = \sum_{i=0}^{s-1} \sum_{j=0}^{s-1} D[r \cdot s + i, c \cdot s + j]
\end{equation}
where $s$ is the output stride (default: 8) and $[r,c]$ index the downsampled grid.

This summation-based downsampling preserves the integral property of the density map: the sum over all pixels equals the total object count. This is a critical property for crowd counting applications, as it ensures that network predictions, when integrated over the image, yield accurate total counts.

\item \textbf{Verification and Quality Control:} The point-to-density conversion is checked for self-consistency by comparing the integral of the density map to the annotated object count:
\begin{equation}
N_{\text{objects}} \approx \sum_{x,y} D(x,y)
\end{equation}
Because each annotated point contributes a unit integral to the density map (one count each), the summation over the density map equals the number of annotated points up to a small residual (typically $< 2\%$) from discretization and from strips whose centroid projects onto the frame boundary. This checks the consistency of the derived density-map targets, not the accuracy of the annotations, which are exact by construction.
\end{itemize}

\subsubsection{Dataset Splitting and Organization}

The complete synthetic dataset is partitioned into training, validation, and optionally test subsets using stratified random sampling. Stratification is performed based on key simulation parameters (e.g., random seed, measurement frame, density level) to ensure balanced representation of different scenarios across splits. A typical split ratio follows the convention of 70\% training, 15\% validation, and 15\% test, though this can be adjusted based on dataset size and application requirements.
For each subset, images and their corresponding annotations are arranged according to the standard Ultralytics YOLO dataset directory structure. Additional annotation types—such as CSV point annotations for density-estimation tasks—follow the same split-based organization. This convention ensures compatibility with established training pipelines and allows seamless integration with tools expecting the standard YOLO dataset layout. Additionally, a YAML configuration file accompanies each dataset, specifying class labels, dataset paths, and metadata for reproducibility.

\subsection{Batch Processing and Reproducibility}

Large-scale synthetic data generation requires efficient batch processing infrastructure and rigorous reproducibility guarantees. The proposed framework implements several mechanisms to ensure scalability, determinism, and full traceability of simulation results.


First, all stochastic elements of the simulation (strip dimensions, spawn positions, initial orientations, and spawn scheduling) are driven by a seeded pseudo-random number generator: Python's \texttt{random} module is seeded once per run with the integer seed from the configuration file, so that an identical configuration and seed reproduce pixel-identical results independently of hardware platform or execution order. This determinism enables exact dataset regeneration from a published configuration and supports controlled parameter studies.

Complete reproducibility further requires a permanent record of every input parameter. A timestamped copy of the JSON configuration is archived alongside the generated data before each batch run, and all critical parameters (simulation settings, geometry, belt speed, random seed, and frame number, together with execution timestamps) are additionally stored in a single SQLite database with the computed density metrics and references to the corresponding images and annotations. This dual storage supports flexible querying by parameter value and automated quality control; the full database schema and on-disk file layout are documented in the released code repository rather than reproduced here.

\subsection{Adherence to FAIR Principles}

The pipeline is designed to align with the interoperability and reuseability aspects of the FAIR data principles \cite{wilkinson_fair_2016}.

All outputs are stored in open, standard formats (PNG images, CSV/TXT annotations, SQLite database, JSON configuration), requiring no proprietary software for access. Additionally, annotation formats conform to established standards (YOLO, CSV point lists) compatible with major deep learning frameworks (PyTorch, TensorFlow, Ultralytics), thus making the pipeline's results highly interoperable.

Since all parameters, deterministic execution, and archived configurations are fully documented, exact reproduction or controlled modification of datasets by independent researchers is supported, making the pipeline also highly reusable.

\subsection{Computational Cost}
\label{sec:compute_cost}

The full dataset was generated on the high-performance compute cluster of RPTU Kaiserslautern-Landau, with each configuration run on a single node providing $8$ CPU cores, $48$~GB of RAM, and one NVIDIA V100 GPU, using Blender~4.0 with the Cycles renderer at a rendering resolution of $512\times512$~px. Table~\ref{tab:runtime} reports the wall-clock cost of the analysis phase (rendering the four analyzer images per measurement frame, computing the density metrics, and writing the annotations), reconstructed from the per-frame database insertion timestamps. The cost per measurement frame grows with the number of deposited strips, from about $11$~s at $1000$ strips to about $26$~s at $2000$ strips (peaking near $27$~s at $1750$ strips), so that rendering and annotating the $33$ measurement frames of a single run takes roughly $6$ to $14$~min. These figures exclude the one-time scene setup and rigid-body physics bake that precede the first measurement frame, which add a smaller, density-dependent overhead per run. Because generation is fully automated and the seeded runs are mutually independent, the corpus scales horizontally across cluster nodes.

\begin{table}[htbp]
\centering
\caption{Wall-clock cost of the generation analysis phase (four polarization renders, density computation, and annotation per measurement frame), by deposited strip count, reconstructed from database timestamps on the hardware above. Each run comprises $33$ measurement frames; each configuration comprises five seeded runs. One-time scene setup and physics baking are excluded.}
\label{tab:runtime}
\begin{tabular}{lcc}
\hline
Strips per run & Time per measurement frame & Time per run ($33$ frames) \\
\hline
$1000$ & $11$~s & $6.0$~min \\
$1250$ & $11$~s & $5.9$~min \\
$1500$ & $14$~s & $7.5$~min \\
$1750$ & $27$~s & $14.3$~min \\
$2000$ & $26$~s & $13.7$~min \\
\hline
\end{tabular}
\end{table}



\section{Results and Conclusions}

To evaluate the performance of the pipeline, a synthetic dataset of over $3{,}300$ measurement frames is created, each rendered in five image components (four analyzer angles plus the derived AoLP map; over $16{,}000$ images in total), of which $3{,}145$ carry complete strip annotations. Here a strong emphasis is set on the data generation methodology, in contrast to exhaustive model performance benchmarking.
The focus is accordingly set on comparison to evaluatable connections known from the real-world system such as spatial characteristics and preliminary validation through deep learning models.

\subsection{Strip-Count Based Density Estimation}
\label{sec:synthetic_dataset}
\begin{figure}[htbp]
    \centering
    \includegraphics[width=\textwidth]{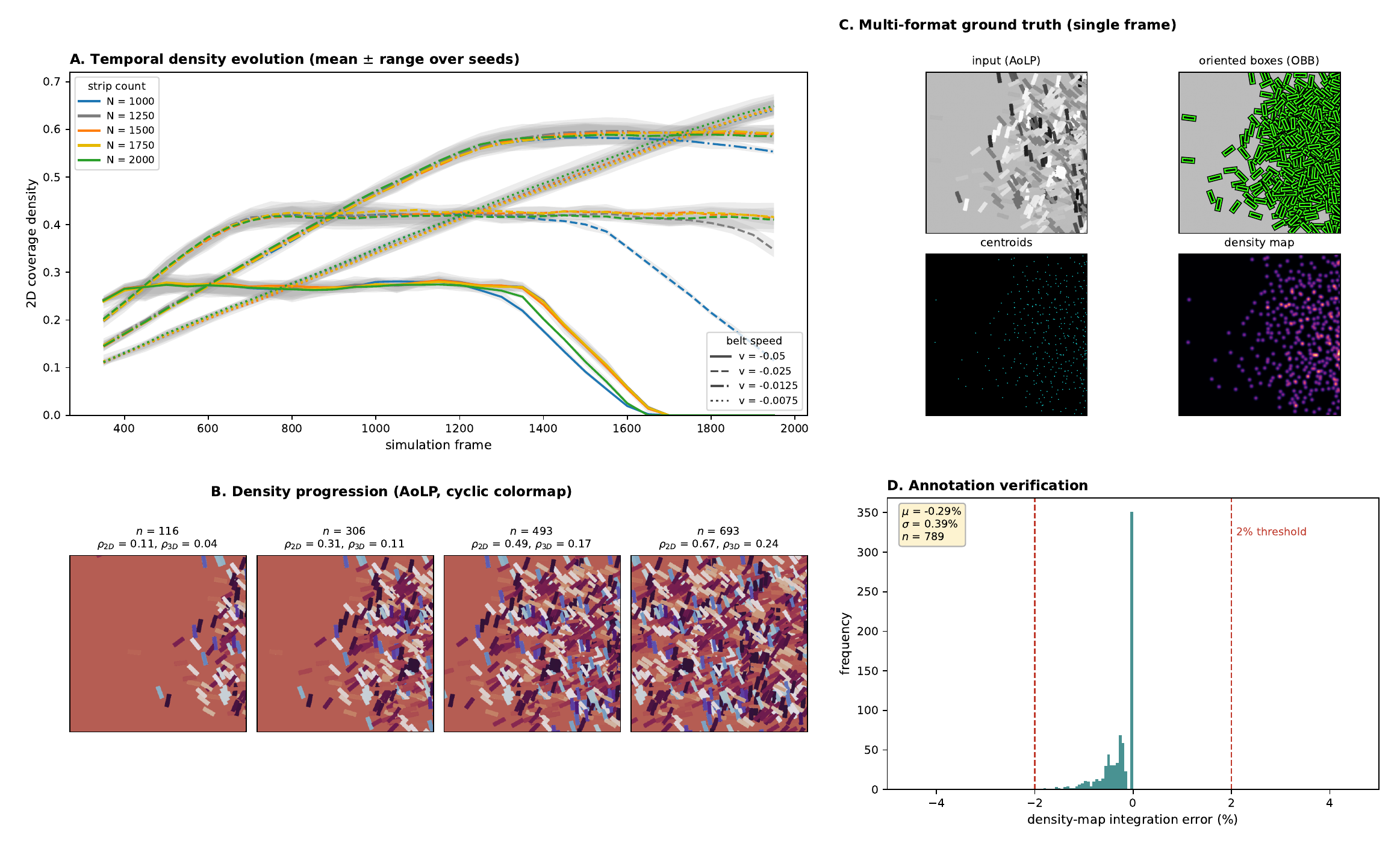}
    \caption{\textbf{Systematic parameter control and multi-format ground truth generation.} 
    (A) Temporal evolution of strip density during the deposition simulation. Different curves represent varying configured strip counts ($N$) and belt speeds ($v$), with shaded regions indicating the mean $\pm$ range across multiple random seeds. Lower belt speeds converge to a steady-state packing density after an initial transient; at higher belt speeds the measured density instead decays toward zero once spawning ends, as the accumulated strips are transported out of the measurement volume. Both behaviors are reproducible across seeds.
    (B) Density progression: representative rendered frames (AoLP, cyclic colormap) at increasing deposited density, each labeled with the number of strips inside the measurement box ($n$) and the 2D and 3D density metrics.
    (C) Multi-format ground truth for a single frame (shown at moderate density for legibility; the label format is identical at all densities), all generated simultaneously from the simulation state: the input AoLP image, the oriented bounding boxes (OBB), the point (centroid) annotations, and the derived density map. The oriented boxes follow each strip's true orientation and are exact and per-strip varying, not axis-aligned or at a constant angle.
    (D) Annotation self-consistency over a representative sample of scenes across all configurations: the density-map integral matches the annotated object count closely (mean $-0.29\%$, essentially all within $2\%$), the small residual reflecting spatial discretization and boundary-projected strip centroids.}
    \label{fig:parameter_control}
\end{figure}

The rigid-body simulation's ability to produce realistic fiber accumulation dynamics is demonstrated through temporal density profiles (Fig.~\ref{fig:parameter_control}A). Density measurements were recorded at multiple frames throughout each simulation, capturing the progression from initial deposition through steady-state packing. For each parameter configuration, simulations were executed across multiple random seeds, with the shaded regions indicating the mean $\pm$ range of observed densities. The curves exhibit characteristic behavior: an initial transient phase where strips progressively accumulate on the belt surface, followed by convergence toward a steady-state density determined by the balance between strip count and belt speed. Higher strip counts ($N$) produce proportionally higher equilibrium densities, while slower belt speeds (smaller $|v|$) allow more strips to accumulate within the measurement region before being transported downstream. The consistency of convergence behavior across random seeds validates both the numerical stability of the physics solver and the effectiveness of the collision-free spawn scheduling algorithm, while the observable variance quantifies the inherent stochasticity of the dropping process.

Areal weight is an important quality parameter of SMC sheets \cite{kraemer_quality_2014, bretz_non-destructive_2021}.
Using identical raw materials, different areal weights can be achieved by varying the relationship of conveyor speed to fiber cutting rate and thus the amount of individual instances on a certain area of the band \cite{orgeas_sheet_2012}.
For real life data, this can easily be evaluated through the 2D areal density (coverage density), which is reflected by the DOLP image component, where fibers and background can be easily separated \cite{duhovic_3d-digitalization_2024, schommer_polarization_2023}.
There, a higher areal weight corresponds to a higher areal density. We expect this relationship to be approximately linear at low areal weights and to approach limited growth at higher weights, since increasing fiber overlap progressively reduces the additional coverage gained per added strip.

This correlation can be replicated by tracking the spawned strip count and comparing it to the calculated coverage, as shown in (Fig.~\ref{fig:parameter_control}A). Strip count was systematically varied over the same area to reproduce different synthetic fiber areal weights, resulting in a corresponding variation of coverage density. The spawn count was swept across five levels ($1000$ to $2000$ strips per run) crossed with four belt speeds. This range was chosen to span the deposition regimes of interest: the lower counts reproduce the sparse and transient accumulation phases, while the higher counts reach several hundred strips within the measurement window, up to the densely overlapping coverage of the real high-density frames, so that the trained models are exercised across the full density range rather than tuned to a single operating point. The controlled exploration of distribution patterns in this way represents the expected correlation between fiber areal weight and coverage density \cite{bretz_non-destructive_2021, duhovic_3d-digitalization_2024}.
(Fig.~\ref{fig:parameter_control}B) shows representative examples spanning from sparse to dense configurations, with their corresponding metrics. They showcase the pipeline's ability to generate a broad range of scenarios whilst automatically calculating metrics reliably. The percentage of visible background color directly represents the areal density and can thus be compared to DOLP images from real processes, although the simulation has not been tuned to the point of allowing quantitative comparisons yet. 


Additionally, the simulation is controlled for consistency by simultaneous verification of the actual spawned volume of strips representing fibers against the number of generated annotations for the deep learning models, as those are essential for high training data quality. (Fig.~\ref{fig:parameter_control}C) showcases the different annotation types, which are generated simultaneously to prevent potential inconsistencies that could arise from sequential conversion processes. A density map is also included, providing through-thickness information more clearly than the annotation maps.

The oriented boxes and point annotations are exact by construction: they are read directly from the simulation state rather than inferred from pixels, so there is no annotation positional error to report. As an internal self-consistency check on the point-to-density conversion, we integrate the Gaussian density map (with mass-conserving boundary handling) over all pixels and compare it to the annotated object count, which it must equal because each annotation contributes a unit integral (one count). Over a representative sample of scenes spanning all configurations (Fig.~\ref{fig:parameter_control}D), the two agree closely (mean $-0.29\%$, standard deviation $0.39\%$), with essentially all scenes within $2\%$; the small residual arises from spatial discretization and from strips whose centroid projects onto the frame boundary, not from any annotation error. This confirms that the point-to-density conversion preserves the object count.

\subsection{Fiber Orientation Distribution}

Fiber orientation distribution is another important parameter in in-line quality control in SMC manufacturing \cite{kraemer_quality_2014, schladitz_non-destructive_2017}.

\begin{figure}[htbp]
    \centering
    \includegraphics[width=\textwidth]{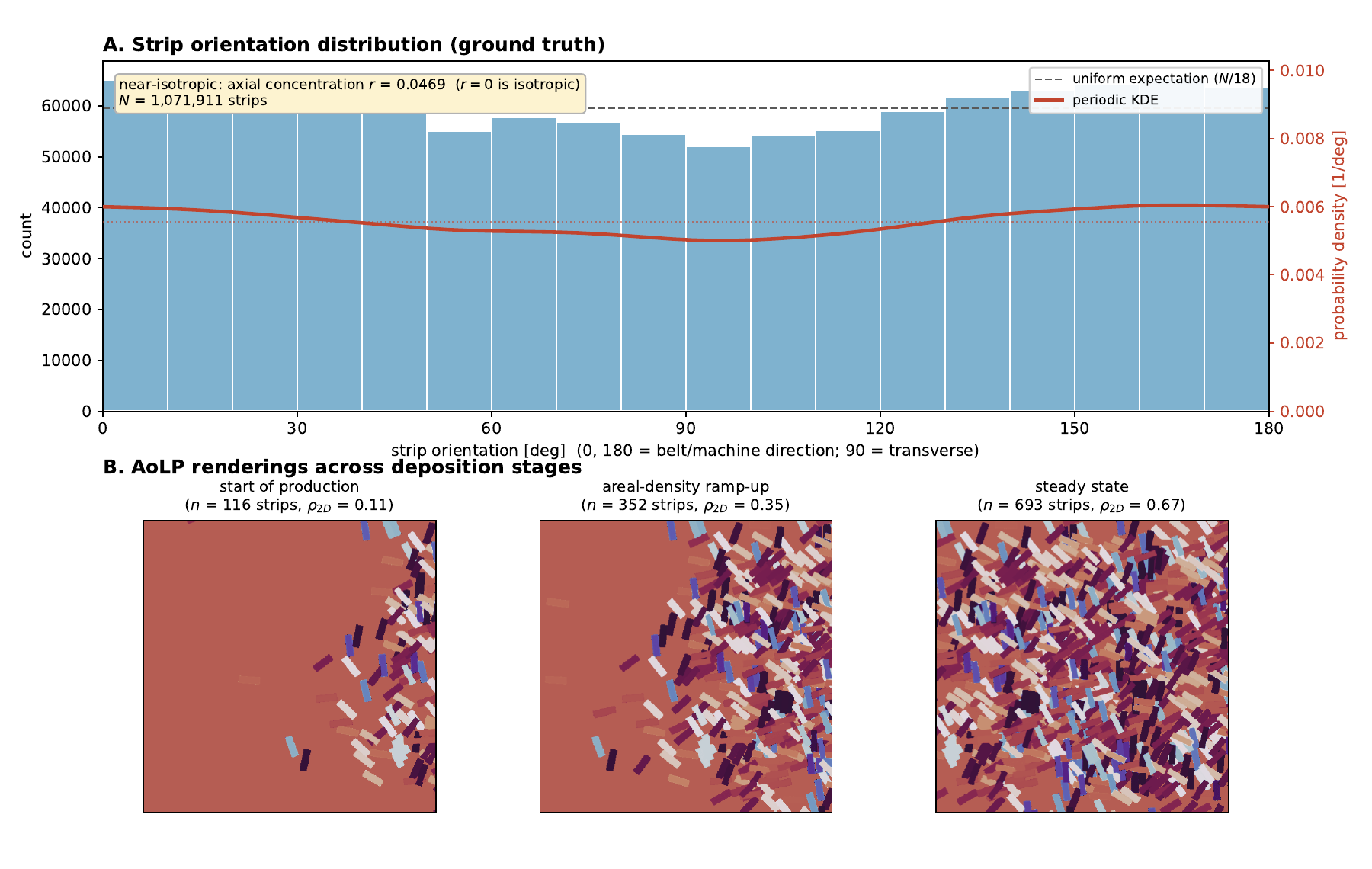}
    \caption{Fiber orientation distribution of synthetic datasets:
    (A) Orientation histogram over the full corpus of $3{,}145$ annotated frames ($1.07$ million labeled strips), with the uniform-expectation line and a periodic kernel density estimate on the secondary density axis; the axial concentration is $r = 0.047$ (statistics in the text), quasi-isotropic up to a weak deposition-induced bias and consistent with as-deposited commercial SMC sheets.
    (B) AoLP renderings across manufacturing stages, capturing orientation information suitable for validating detection algorithms and training orientation-sensitive networks.}
    \label{fig:spatial_characteristics}
\end{figure}

Fig.~\ref{fig:spatial_characteristics}A shows the numeric fiber orientation distribution of the synthetic dataset. Across all $3{,}145$ annotated frames ($1{,}071{,}911$ strips), the distribution is close to uniform: the axial circular concentration is $r = 0.047$ (where $r = 0$ denotes perfect isotropy), and no $10^\circ$ orientation bin deviates from the uniform expectation by more than $13\%$. At this sample size, a chi-square goodness-of-fit test \cite{pearson1900,rice2007} nevertheless rejects exact uniformity ($\chi^2 = 5119$, $17$ degrees of freedom), reflecting a weak orientation preference introduced by the deposition dynamics. The same statistics computed separately on the training pool and the held-out test split closely match ($r = 0.047$ and $r = 0.053$, respectively), so the near-isotropy is a property of the generated data as a whole rather than of a particular split. The weak departure from uniformity is directional: the excess falls in the orientation bins near $0^\circ$ and $180^\circ$ and the deficit near $90^\circ$, with the largest bins about $9\%$ above and the smallest about $13\%$ below the uniform expectation. This is a slight preference for the belt (machine) direction over the transverse direction, the expected signature of directional deposition and belt transport. A directional preference of the same sign has been reported for real carbon-fiber SMC, where roving alignment along the production direction was observed by combined polarization imaging and X-ray computed tomography \cite{duhovic_quantitative_2025}; the synthetic bias found here is far weaker, consistent with the near-isotropic as-deposited regime under study. Its magnitude is small (all bins within $13\%$ of uniform, $r = 0.047$), so its influence on bulk mechanical properties is minor; quantifying the corresponding orientation bias on real material is a target of future in-line evaluation.
This near-isotropic state is consistent with the planar quasi-isotropic fiber orientation reported for as-deposited, randomly oriented commercial SMC, prior to the flow-induced reorientation that occurs during compression molding, and thus demonstrates the simulation's ability to reproduce this property of the freshly deposited material \cite{schladitz_non-destructive_2017, feraboli_characterization_2009, oldenbo_global_2004}.

The AoLP visualization examples in Fig.~\ref{fig:spatial_characteristics}B demonstrate that orientation information is accurately captured in rendered images, making them suitable as surrogate data for actual camera images. This holds across different stages of sheet manufacturing, from the start of production through the ramp-up of areal density to steady-state manufacturing. The rendered orientation signal is equally reliable at partial and at high coverage, so it is not limited by the fact that the present simulations do not yet reach the nominal areal density of one. Consequently, these polarization-based renderings provide valuable ground truth for validating orientation detection algorithms and can serve as additional training signals for orientation-sensitive networks. The ability to generate synthetic polarization data addresses a gap in available training datasets for polarization-based industrial inspection systems \cite{hutten_deep_2024}.

\subsection{Model Validation on Synthetic Data}
\label{sec:synthetic_validation}

To verify that the generated data can train deep-learning models to convergence, two representative architectures were trained exclusively on synthetic data: YOLOv11 \cite{khanam_yolov11_2024} (Ultralytics implementation) for single-class strip detection, and CSRNet \cite{li_csrnet_2018} as a density-map baseline for count estimation. This subsection reports their performance on held-out synthetic data; the transfer of these models to images from real SMC manufacturing is assessed in Section~\ref{sec:real_world_transfer}.

For detection, all five YOLOv11 model scales were trained for $100$ epochs on $576$ rendered images with a held-out validation set of $144$ images (single class, \emph{strip}). Table~\ref{tab:yolo_sweep} summarizes the converged validation metrics. All scales reach a precision above $0.98$; the mean average precision grows moderately with model capacity and saturates near $\text{mAP}_{50} = 0.91$ ($\text{mAP}_{50\text{--}95} = 0.90$) for the largest model. Recall stabilizes near $0.82$ across scales, consistent with ground-truth labels that, being generated from the simulation state, include heavily occluded strips that no detector can fully recover from a single image.

\begin{table}[htbp]
\centering
\caption{YOLOv11 strip detection on the held-out synthetic validation set ($144$ images) after $100$ converged training epochs, across the five model scales.}
\label{tab:yolo_sweep}
\begin{tabular}{lcccc}
\hline
Model & Precision & Recall & $\text{mAP}_{50}$ & $\text{mAP}_{50\text{--}95}$ \\
\hline
YOLOv11n & 0.984 & 0.793 & 0.883 & 0.812 \\
YOLOv11s & 0.995 & 0.810 & 0.896 & 0.859 \\
YOLOv11m & 0.997 & 0.818 & 0.903 & 0.884 \\
YOLOv11l & 0.997 & 0.820 & 0.902 & 0.887 \\
YOLOv11x & 0.998 & 0.822 & 0.908 & 0.899 \\
\hline
\end{tabular}
\end{table}

\begin{figure}[htbp]
    \centering
    \includegraphics[width=\textwidth]{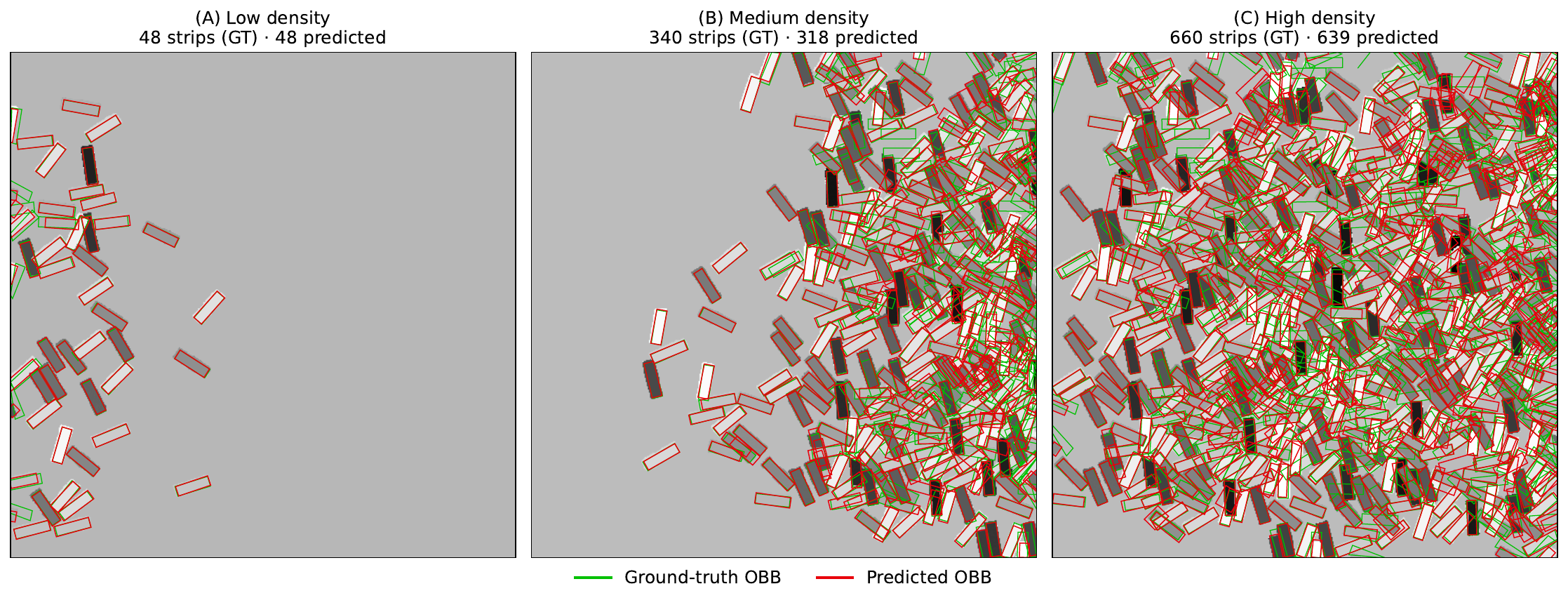}
    \caption{\textbf{Oriented-bounding-box detection across density regimes.} Ground-truth (green) and predicted (red) oriented bounding boxes from the single-scale YOLO11s-OBB detector on three held-out synthetic validation frames spanning the density range: (A) low ($48$ strips), (B) medium ($340$ strips), and (C) high ($660$ strips). Predictions are shown at a confidence threshold of $0.25$ with the per-image detection cap raised to accommodate the densest frames. The predicted boxes follow each strip's position, size, and orientation, and their count closely tracks the ground truth even under the heavy occlusion of the high-density regime.}
    \label{fig:obb_detection}
\end{figure}

The YOLOv11 sweep above uses the axis-aligned detection task, whereas the annotation format that the digital twin emits and that this paper foregrounds is the oriented bounding box. To confirm that the exact OBB labels support oriented detection directly, we additionally trained a single-scale YOLO11s-OBB detector on the AoLP renders. It was evaluated on a leakage-safe split in which one random seed per configuration is held out for validation: whole simulation runs are assigned to either training or validation and never split within a run, so that near-duplicate frames from the same run cannot leak across the split ($440$ training and $220$ validation frames, each containing between $48$ and $660$ strips). Trained as a single-scale proof of concept ($15$ epochs on a single Apple-silicon GPU using the Metal Performance Shaders backend), the oriented detector reaches $\text{mAP}_{50} = 0.83$ and $\text{mAP}_{50\text{--}95} = 0.72$ (precision $0.97$, recall $0.71$) at the default cap of $300$ detections per image, directly comparable to the axis-aligned sweep in Table~\ref{tab:yolo_sweep}; raising the cap to accommodate the densest frames, which contain up to $660$ strips, recovers recall to $0.80$ ($\text{mAP}_{50} = 0.89$). The predicted oriented boxes follow the ground-truth strips in position, size, and orientation across the full density range, and the number of predicted boxes closely tracks the ground-truth strip count from the sparsest to the densest frames ($48$ versus $48$, $340$ versus $318$, and $660$ versus $639$ strips for the representative frames in Fig.~\ref{fig:obb_detection}). A full multi-scale OBB sweep trained to convergence is left to future work; this single-scale result already establishes that the oriented annotations are directly trainable, so that the reported detector matches the oriented-box format used throughout the paper.

For count estimation, CSRNet was trained on the Gaussian density-map targets described in Section~\ref{sec:density_map_generation}, using a seeded split of $300$ training, $100$ validation, and $100$ test frames drawn evenly across the generation runs ($25$ frames per run); the checkpoint with the best validation error was used for evaluation. Table~\ref{tab:csrnet} reports the count error on the $100$-frame test split, overall and per density regime, where the regimes are empirical tertiles of the ground-truth strip count (low $\le 220$, medium $221$--$375$, high $> 375$ strips per frame). CSRNet is included as a widely used density baseline rather than as a proposed solution. Its overall MAE of $121.3$ strips stems from a strongly compressed prediction range: trained on this density sweep, it regresses toward the mid-density mode, with predictions spanning only $298$--$377$ strips against ground-truth counts of $0$--$633$, over-predicting sparse frames and under-predicting dense ones, so it is numerically accurate only in the mid-density regime. The wide density sweep of the dataset thus exposes a failure mode of density-map counting that a benchmark confined to the typical density range of crowd-counting datasets would not reveal.

\begin{table}[htbp]
\centering
\caption{CSRNet count error on the held-out synthetic test split ($100$ frames). Density regimes are empirical tertiles of the ground-truth strip count per frame.}
\label{tab:csrnet}
\begin{tabular}{lccc}
\hline
Density regime & $n$ & MAE & RMSE \\
\hline
low ($\le 220$) & 32 & 189.7 & 202.0 \\
medium ($221$--$375$) & 39 & 42.3 & 55.1 \\
high ($> 375$) & 29 & 151.9 & 166.3 \\
\hline
overall & 100 & 121.3 & 149.2 \\
\hline
\end{tabular}
\end{table}

\subsection{Synthetic-to-Real Appearance Fidelity and Preliminary Model Validation}
\label{sec:real_world_transfer}

With the two synthetic-trained models from Section~\ref{sec:synthetic_validation} in hand, we assess their transfer to images from real SMC manufacturing. We state at the outset what can and cannot be established on real data at this material's overlap density. A real strip-deposition frame cannot be inverted to the digital-twin parameters that would have produced it; this impossibility is itself part of the motivation for a simulation that emits exact labels. We therefore neither claim a paired, pixel-matched synthetic-to-real scene, nor attempt an exhaustive per-strip real ground truth, which, as shown below, cannot be produced by hand at several hundred overlapping strips per frame. Instead we characterize transfer along four complementary axes that require no real labels: (i) qualitative appearance fidelity, (ii) unpaired distributional fidelity between the synthetic and real image sets, (iii) a small human-counted probe as an order-of-magnitude anchor, and (iv) a label-free temporal-consistency proxy on real video.

\begin{figure}[htbp]
    \centering
    \includegraphics[width=\textwidth]{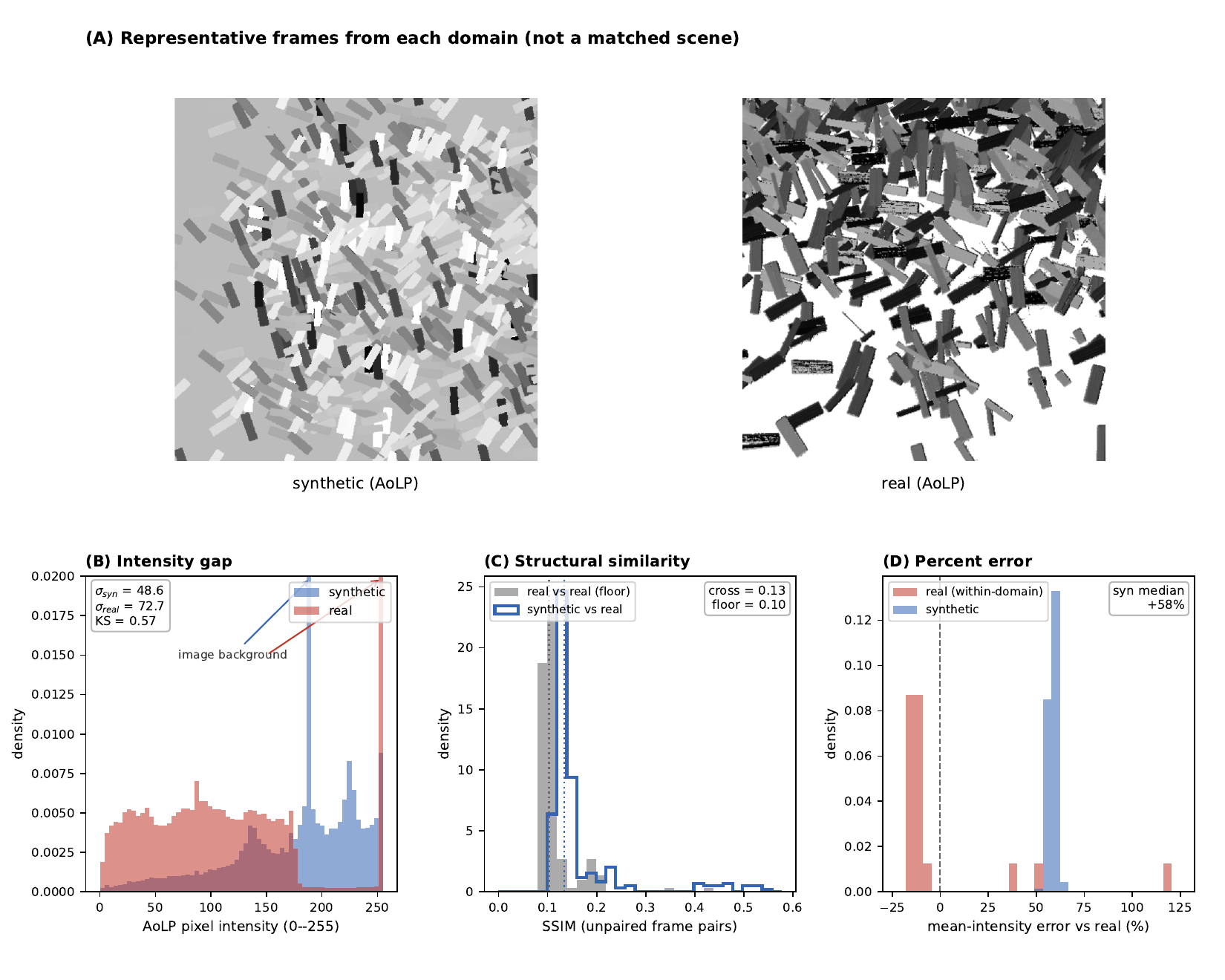}
    \caption{\textbf{Synthetic-to-real domain gap.}
    (A) Representative synthetic (left) and real (right) frames; the panel illustrates similar overall geometry and orientation texture, not a matched scene. The visible residuals are background appearance and slightly sharper synthetic edges, quantified in Fig.~\ref{fig:appearance_fidelity}A.
    (B) Per-pixel AoLP intensity distributions of the two domains (standard deviations $\sigma_{\text{syn}} = 48.6$ and $\sigma_{\text{real}} = 72.7$ on the $0$--$255$ scale; $\text{KS} = 0.57$). As analyzed in the text, this raw-intensity difference reflects the difference in background type (a bare conveyor belt rendered at a near-constant polarization value versus the resin- and fiber-covered real background) together with a reference-axis calibration offset in the polarization channel, rather than a difference in strip appearance.
    (C) Structural similarity (SSIM) over unpaired frame pairs: the synthetic-to-real distribution (median $0.13$) is comparable to the real-vs-real reference floor (median $0.10$), because the strips occupy uncorrelated positions and pixel-aligned similarity is therefore intrinsically low even between two real frames.
    (D) Per-frame mean-intensity percent error relative to the real-set mean; the synthetic frames cluster at a systematic offset of $\approx +58\%$ (a background-type and polarization-calibration difference), against the broader frame-to-frame spread of the real set.}
    \label{fig:real_world_transfer}
\end{figure}

At the qualitative level (Fig.~\ref{fig:real_world_transfer}A), a representative synthetic frame and a representative real frame are alike in strip geometry and orientation texture. This is a representative-appearance comparison and not a matched pair: as noted above, the specific real configuration cannot be reproduced synthetically, so the panel shows that the two domains look alike, not that a given real scene has been recreated.


The raw-appearance difference between the two domains is quantified along three complementary axes in Fig.~\ref{fig:real_world_transfer}B--D, computed on the un-warped, density-matched frame sets defined below. The per-pixel AoLP intensity distributions differ mainly in spread and mean (standard deviations $\sigma_{\text{syn}} = 48.6$ versus $\sigma_{\text{real}} = 72.7$ on the $0$--$255$ scale, and a per-frame mean offset of about $+58\%$, Fig.~\ref{fig:real_world_transfer}D); as analysed in the following paragraph, this absolute-value gap stems from the difference in background type between the simulated bare belt and the resin- and fiber-covered real line, together with a polarization reference-axis calibration offset, rather than from the rendered strip appearance. Because the strips occupy uncorrelated positions from frame to frame, a pixel-aligned structural-similarity index (SSIM) is intrinsically low even between two distinct real frames (median $0.10$). The synthetic-to-real SSIM (median $0.13$) is comparable to, and marginally above, this within-real reference floor (Fig.~\ref{fig:real_world_transfer}C), while remaining well below the within-synthetic self-similarity (median $0.26$); the cross-domain similarity is therefore on par with the intrinsic frame-to-frame similarity of the real process and is not inflated by the uniform synthetic background, an ordering preserved by a background-invariant Sobel-gradient SSIM (cross-domain $0.09$ versus a within-real floor of $0.08$).

\begin{figure}[htbp]
    \centering
    \includegraphics[width=\textwidth]{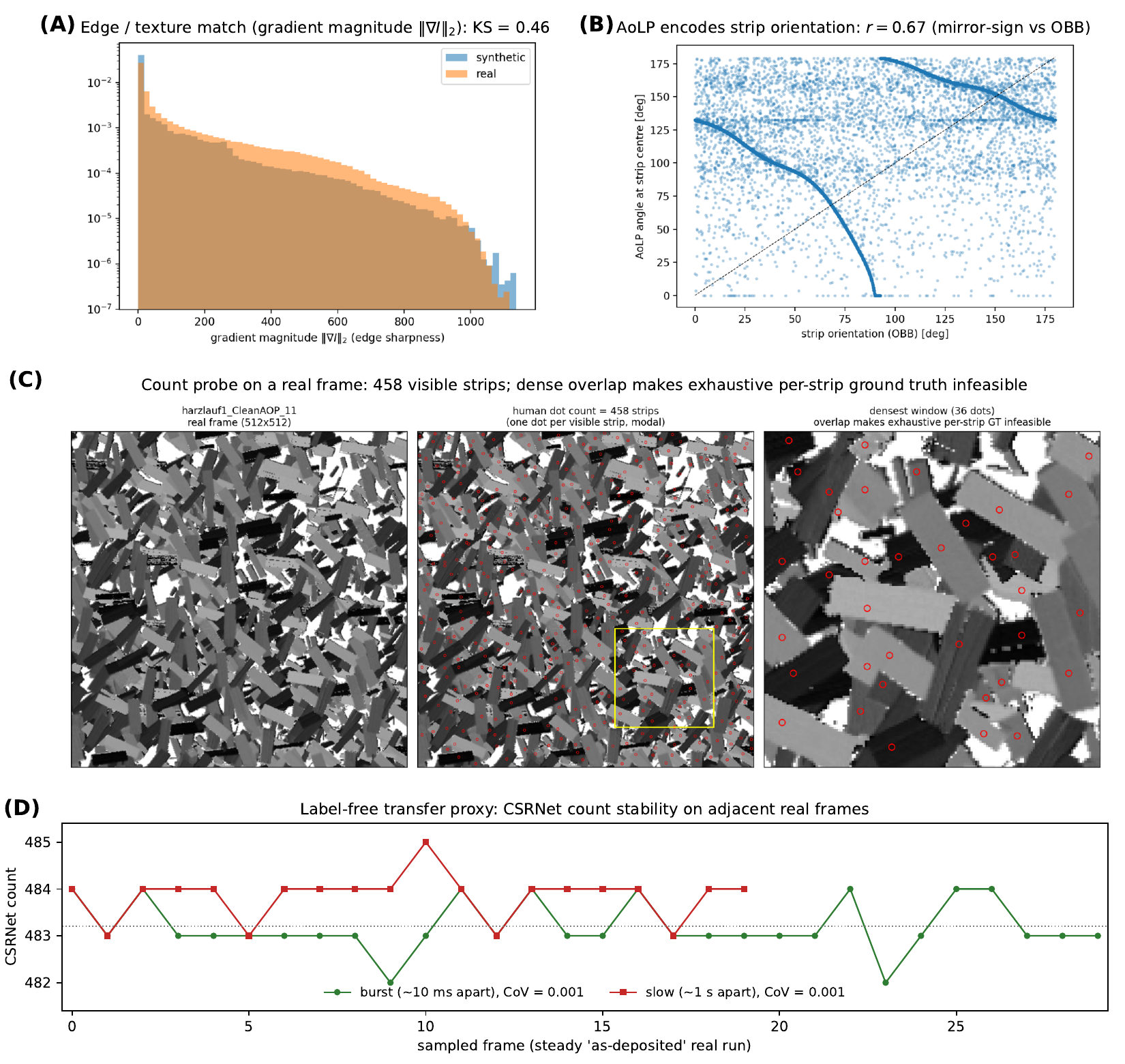}
    \caption{\textbf{Label-free synthetic-to-real fidelity and transfer evidence.}
    (A) Unpaired distribution of the per-pixel Sobel gradient magnitude (edge and texture sharpness) for the synthetic and real image sets; the two distributions overlap by $0.56$ ($\text{KS} = 0.46$), matching the overlap of $0.57$ measured between two real acquisition series of the same material (Table~\ref{tab:ks_reference}).
    (B) Rendered AoLP sampled at strip centroids versus the ground-truth oriented-box angle; the mirror-symmetric trend ($r = 0.67$) shows that the polarization channel encodes strip orientation.
    (C) A real frame dot-counted by hand (one dot per visible strip): $458$ strips in a $512\times512$ crop, with a densest-window inset showing that occlusion makes exhaustive per-strip ground truth infeasible.
    (D) Temporal stability of the synthetic-trained CSRNet count across consecutive and strided frames of a steady real run (coefficient of variation $0.001$), a label-free consistency check.}
    \label{fig:appearance_fidelity}
\end{figure}

Because it needs neither pairing nor ground truth, the quantitative backbone of the evaluation is an unpaired comparison of appearance statistics computed over the synthetic image set against the real image set, density-matched to the high-coverage regime that the real frames occupy: the synthetic set comprises $150$ AoLP renders whose simulated strip count lies in $[450, 1000]$, and the real set comprises the $18$ highest-coverage of the $23$ available discrete-strip laboratory frames (the five sparsest, which contain too few strips for the comparison, are excluded). Edge and texture structure, quantified by the per-pixel Sobel gradient magnitude, gives a Kolmogorov--Smirnov distance of $\text{KS} = 0.46$ between the two distributions, corresponding to a histogram overlap of $0.56$ (Fig.~\ref{fig:appearance_fidelity}A); this is the smallest distance of the compared groups (a mismatched production-texture set gives $\text{KS} = 0.66$, overlap $0.35$), and the synthetic edges are only marginally sharper in the extreme tail. The Kolmogorov--Smirnov distance is simply the largest vertical gap between the two empirical cumulative distribution functions; it describes how far apart two distributions lie and is neither a correlation nor a score with an intrinsic scale. It accordingly has no sample-size-independent acceptance threshold; the critical value of the distance shrinks with the two sample sizes as
\begin{equation}
\label{eq:ks_critical}
D_{\alpha} = c(\alpha)\sqrt{\frac{1}{n} + \frac{1}{m}},
\end{equation}
where $n$ and $m$ are the sizes of the two samples and $c(\alpha)$ is the tabulated coefficient of the limiting distribution, $c = 1.36$ at the $5\%$ level \cite{massey1951}. For the pixel samples compared here Eq.~(\ref{eq:ks_critical}) is of order $10^{-3}$, and it remains orders of magnitude below every distance reported in this section even after allowing for the spatial correlation between neighbouring pixels, which reduces the effective sample size well below the nominal pixel count. Any two non-identical distributions are therefore separated at conventional significance, and statistical significance carries no information about fidelity here. The same sample-size sensitivity is documented for the distribution-level appearance metrics in common use for comparing image sets, whose values are biased and shift with the number of samples to the point of reordering the models being compared \cite{chong2020, jayasumana2024}; such distances are consequently read against a reference measured under an identical protocol rather than against a fixed bound. We accordingly interpret ours against within- and between-domain references computed with the identical statistic, pixel sample, and histogram grid (Table~\ref{tab:ks_reference}). Two disjoint halves of a single real run differ by $\text{KS} = 0.03$ (rising to $0.09$ for a contiguous rather than an interleaved split), the repeatability floor under unchanged acquisition conditions, whereas two different real acquisition series of this same material, the discrete-strip laboratory run and the production run, differ by $\text{KS} = 0.41$ (overlap $0.57$). The synthetic-to-real distance of $0.46$ (overlap $0.56$) is thus of the same magnitude as the distance between two real recordings of the material: on this statistic the separation is dominated by acquisition conditions and scene type rather than by the synthetic origin of the images. Part of the residual distance is expected rather than a sign of mismatch: the perfectly flat synthetic surfaces drive the gradient magnitude toward zero in homogeneous regions, whereas real sensor noise produces small non-zero gradients even where the surface is uniform, so the two low-gradient tails cannot coincide exactly (consistent with Fig.~\ref{fig:appearance_fidelity}A). That the comparison is distributional rather than spatially registered is not a limitation here, because the strips occupy uncorrelated positions between frames, so a pixel-aligned edge comparison would be uninformative, and the two distributions agree most closely in the high-gradient edge range that carries the texture signal. The angle-of-linear-polarization channel faithfully encodes strip orientation: sampled at strip centroids across $40$ frames ($n = 21{,}530$ strips), the rendered AoLP correlates with the ground-truth oriented-box angle at $r = 0.67$ (Fig.~\ref{fig:appearance_fidelity}B), confirming that the polarization rendering carries the physical orientation signal rather than a constant or artifactual value. The synthetic strip orientations are themselves quasi-isotropic (orientation concentration $r = 0.06$), consistent with the fiber-orientation distribution reported above (Fig.~\ref{fig:spatial_characteristics}A). The remaining distributional discrepancies are characterized rather than attributed to strip appearance. The absolute AoLP value differs (Fig.~\ref{fig:real_world_transfer}B; $\text{KS} = 0.57$) largely because the polarization camera's zero reference is its physical mounting angle, unrelated to the simulator's coordinate frame: applying a single $120^\circ$ reference-axis offset raises the AoLP distribution overlap from $0.31$ to $0.60$, so a single calibration constant closes about half of the value gap and brings the cross-domain overlap to the level found between the two real acquisition series, whose AoLP distributions overlap one another by $0.66$. The real AoLP images additionally follow the Baumer camera's standard false-color mapping, in which orientation is quantized into fixed one-degree intensity bins and every pixel below a degree-of-polarization threshold is set to the maximum value ($255$, rendered white) rather than spanning the full available range; this convention concentrates real values near the top of the intensity scale, visible in Fig.~\ref{fig:real_world_transfer}B, and is not reproduced by the linear synthetic mapping, which accounts for part of the absolute-value gap. The larger discrepancies in per-image mean and foreground-coverage fraction ($\text{KS} \approx 0.94$; nominal Otsu coverage $\approx 0.79$ for synthetic versus $\approx 0.48$ for the real set) reflect the same background-type and calibration differences rather than a difference in strip content. These last two figures are per-image statistics over the $18$ real frames rather than pixel-level distributions, so the statistic is quantized in steps of $1/18$ and is not on the same footing as the pixel-level distances above. This foreground statistic is moreover not directly comparable across the two domains, because Otsu thresholding selects the bright bare belt in the synthetic frames but the bright strips in the real frames; and the simulated bare conveyor belt is deliberately not intended to reproduce the resin- and fiber-covered real background. These are calibration and background effects rather than failures of the rendered strip appearance, whose edge and orientation structure matches.

\begin{table}[htbp]
\centering
\caption{Reference values for the unpaired appearance comparison, all computed on the per-pixel Sobel gradient magnitude with the identical statistic, pixel sample, and histogram grid. The Kolmogorov--Smirnov distance has no sample-size-independent acceptance threshold and is interpretable only against such references: the first two rows bound the distance attainable within one domain, and the third gives the distance between two real acquisition series of the same material.}
\label{tab:ks_reference}
\begin{tabular}{lccc}
\hline
Comparison & KS & Wasserstein & Overlap \\
\hline
Synthetic vs.\ synthetic (two disjoint halves) & 0.01 & 0.8 & 0.99 \\
Real vs.\ real (two disjoint halves of one run) & 0.03 & 11.5 & 0.97 \\
Real laboratory vs.\ real production run & 0.41 & 59.0 & 0.57 \\
Synthetic vs.\ real laboratory & 0.46 & 47.7 & 0.56 \\
Synthetic vs.\ real production (mismatched scene type) & 0.66 & 96.4 & 0.35 \\
\hline
\end{tabular}
\end{table}

The count probe provides an order-of-magnitude anchor on real data. Two real frames were exhaustively dot-counted by hand, counting a strip if any part of it is visible, giving $244$ and $458$ strips in a single $512\times512$ crop (Fig.~\ref{fig:appearance_fidelity}C). We deliberately keep this probe small. At this overlap the manual count is a lower bound with an estimated $10$--$20\%$ systematic undercount, because fully occluded strips cannot be seen (Fig.~\ref{fig:appearance_fidelity}C, inset); the error is systematic rather than random, so it does not diminish by averaging over more frames and it places a hard floor on any count-based real metric. This limitation is itself the central evidence for the approach: an exhaustive, amodal per-strip ground truth for several hundred densely overlapping strips cannot be obtained by manual annotation with any tool, which is precisely the labeling barrier that a simulator emitting exact labels removes.

As a label-free proxy for transfer, we examined the temporal consistency of the synthetic-trained CSRNet count on a steady real video sequence (Fig.~\ref{fig:appearance_fidelity}D). Across $30$ consecutive frames (about $10$~ms apart) and a further $20$ frames strided across the run (about $1$~s apart, $50$ frames in total), the predicted count stays between $482$ and $485$, a coefficient of variation of $0.001$: the model responds smoothly to real input rather than producing erratic or collapsed predictions. Because a model insensitive to its input would also be stable, we present this within-run stability as a label-free consistency check rather than a measure of accuracy; between-run discrimination on runs of differing density is a stronger test left to future multi-run evaluation.

\paragraph{Closing the residual appearance gap}
The one clearly visible appearance gap is that synthetic strip edges are cleaner than real edges, which carry fiber fray, defocus, and sensor noise. This is a standard and well-characterized target for data augmentation applied to the synthetic training data \cite{buslaev_albumentations_2020}: photometric and sensor-noise transformations (Gaussian, Poisson, and sensor noise; defocus and motion blur; brightness, contrast, and gamma jitter) together with edge-specific transformations (morphological erosion and dilation, elastic deformation, and feathered strip boundaries) reproduce these effects, and, more broadly, domain randomization over such appearance factors \cite{tobin_domain_2017} and learned domain adaptation (for example Fourier-domain adaptation \cite{yang_fda_2020}, unpaired image-to-image translation \cite{zhu_unpaired_2017}, or adaptive-instance-normalization style transfer \cite{huang_arbitrary_2017}) provide further routes should augmentation prove insufficient \cite{wang_deep_2018}.
Because the digital twin is fully automated, such refinements can be applied and the training set regenerated rapidly, at a fraction of the cost of collecting and labeling additional real data, turning real-world deployment from a one-time validation step into an iterative improvement loop.

\paragraph{Comparison to Traditional Approaches}
These observations are consistent with the broader synthetic-to-real literature: synthetic data provides a strong initialization for perception models, with domain adaptation or a limited quantity of real labels closing the residual gap \cite{tremblay_training_2018, wang_deep_2018, hinterstoisser_pre-trained_2019}. The distinctive point for this material is that exact labels for the densely overlapping-strip regime are obtainable only in simulation, so bootstrapping model development from synthetic data is not merely a convenience but the only scalable route to labeled training data at this scale.

\subsection{Answers to Research Questions}
This work addresses all three research questions through the development and validation of a process-informed, photorealistic digital twin for C-SMC strip deposition. Regarding the first research question on generating large-scale, fully annotated image datasets \emph{in silico}, the developed simulation framework demonstrates the capability to produce over $16{,}000$ images across more than $3{,}300$ measurement frames with comprehensive annotations, capturing meso- and macro-scale fiber-arrangement variability through rigid-body strip deposition dynamics that incorporate gravitational settling, friction, and belt motion; this variability is defined measurably, through the orientation-concentration and coverage statistics reported in Section~5, rather than asserted qualitatively. The framework generates multiple synchronized annotation formats, including YOLO oriented bounding boxes, point annotations, density maps, and polarization-based renderings (AoLP and DoLP images), while maintaining strict reproducibility through seeded random number generation; the annotations are exact by construction, and the derived density-map targets pass a density-integral self-consistency check to within $2\%$. For the second research question on supporting camera-based quality assurance methods, the results demonstrate that the synthetic datasets successfully replicate key industrial characteristics: temporal density evolution profiles show realistic strip accumulation and steady-state packing behavior matching the balance between strip count and belt speed, while fiber orientation distribution analysis confirms near-isotropic behavior consistent with commercial SMC sheets (axial orientation concentration $r = 0.047$ for the dataset as a whole, with training pool and held-out test split closely matching). The AoLP renderings accurately capture orientation information across different manufacturing stages, providing suitable ground truth for validating detection algorithms and training orientation-sensitive networks. Addressing the third research question on machine learning training and validation, YOLOv11 for object detection and CSRNet for density estimation were trained exclusively on synthetic data and benchmarked on held-out synthetic splits (detection $\text{mAP}_{50}$ up to $0.91$ across model scales, with a dedicated oriented-bounding-box detector confirming that the oriented labels are themselves directly trainable at $\text{mAP}_{50} = 0.83$; count estimation strongest in the mid-density regime); on real manufacturing images, the label-free distributional distance between synthetic and real frames in edge and texture structure is of the same magnitude as the distance between two real acquisition series of the same material (Table~\ref{tab:ks_reference}), and the synthetic-trained density model produces temporally consistent, non-collapsing count predictions on a steady real run. The residual synthetic-to-real gap is characterized rather than assumed away (a polarization calibration offset, a denser-than-real coverage regime, and unmodelled background), and the automated generation pipeline enables the rapid, augmentation-based refinement described in Section~\ref{sec:real_world_transfer}, substantially reducing the labeling burden compared to manual annotation while maintaining the ability to bootstrap model development entirely from synthetic data.


\section{Outlook}
\label{sec:outlook}
Future work will focus on enhancing the realism and utility of synthetic data for fiber orientation analysis and anomaly detection in SMC inspection. One key area is improving the representation of fiber orientations, particularly capturing the non-uniform distributions observed in real-world scenarios, which are critical for robust anomaly detection. Additionally, boundary conditions in synthetic datasets will be refined to better reflect operational realities. For example, training data should include strips partially exiting the measurement volume, in line with the sliding window inference paradigm used during deployment. This can be achieved through adjustments to camera framing or by expanding the spawn volume beyond the nominal measurement box.

Another important direction is increasing lighting diversity. The synthetic data generation pipeline will be expanded to include directional illumination at varying angles (0°–45° from vertical), a wider range of intensities (50–200\% of baseline), and both hard and soft shadows, to better capture real-world variability. Similarly, material appearance will be refined to reduce the gap with real imagery. Adjustments may include increased surface roughness, moderate variations in grayscale tone, and enhanced fiber texture detail using normal mapping to simulate surface irregularities.

Background variation will also be addressed by incorporating diverse textures and colors representative of different conveyor belt materials, such as gray, blue, or black surfaces with smooth or textured finishes, rather than the current uniform gray background. Finally, extreme density scenarios will be more heavily represented in the training data. The strip counts studied here were chosen for the degree of strip overlap they induce rather than for the count alone; higher counts additionally raise the incidence of directly adjacent, equally oriented rovings, which merge visually and form the hardest case for both detection and orientation estimation. Including cases with very high strip counts (>900) will better prepare models for these worst-case real-world conditions, as current datasets emphasize moderate densities for computational efficiency.

Together, these planned improvements aim to increase the realism of synthetic training data and, in turn, enhance the reliability and performance of machine learning models for fiber orientation analysis and anomaly detection in SMC applications.

\section*{Code and Data Availability}
\label{sec:code_availability}
The complete digital-twin pipeline, comprising the rigid-body deposition simulation, the polarization rendering model, the multi-format annotation generation, and the training and evaluation code for the YOLOv11 and YOLO11-OBB detectors and the CSRNet density model, is released as open-source software under the Apache License 2.0. The repository is hosted on GitHub (\texttt{github.com/PatRuediger/c-smc-digital-twin}), and a versioned, archived snapshot will be deposited on Zenodo, with the DOI provided for the camera-ready version. It documents the configuration schema, the SQLite database schema and on-disk file layout, and the tunable physics and appearance parameters (friction, restitution, damping, and the polarization and lighting settings), and includes a small example dataset of representative rendered frames together with their oriented-box, point, and density-map annotations, sufficient to run the pipeline end to end and to regenerate a representative figure. The full generated corpus (over $3{,}300$ rendered measurement frames, of which $3{,}145$ carry complete annotations) is available from the authors on reasonable request.

\section*{Acknowledgements}
The project “Machine Vision Carbon Composites – Digitalization for the Sustainable Manufacturing of Carbon Composites through Polarization Imaging for Quality Assurance and Real-Time Process Control” is funded within the framework of the “Industrial Collective Research” program by the Federal Ministry for Economic Affairs and Climate Action on the basis of a decision by the German Bundestag.





\section*{Declaration of generative AI and AI-assisted technologies in the manuscript preparation process}
During the preparation of this work, the authors used Claude (Anthropic) in order to revise sections of the manuscript text, to improve readability and language, and to assist with reference retrieval and verification. After using this tool, the authors reviewed and edited the content as needed and take full responsibility for the content of the published article.

\bibliographystyle{ieeetr} 
\bibliography{Composites_AI_paper}






\end{document}